\documentclass[letterpaper, 10 pt, conference]{ieeeconf}  

\IEEEoverridecommandlockouts                  

\usepackage{amsfonts}
\usepackage{listings}
\usepackage{xcolor}
\usepackage{booktabs}
\usepackage{threeparttable}

\definecolor{myblue}{rgb}{0.0, 0.0, 1.0}
\definecolor{mygreen}{rgb}{0.0, 0.5, 0.0}
\definecolor{myorange}{rgb}{1.0, 0.647, 0.0}
\definecolor{myred}{rgb}{1.0, 0.0, 0.0}

\definecolor{darkred}{rgb}{0.5, 0.0, 0.0}

\title{\LARGE \bf Towards Data-Driven Metrics\\ for Social Robot Navigation Benchmarking}

\author{Pilar Bachiller-Burgos, Ulysses Bernardet, Luis V. Calderita, Pranup Chhetri, Anthony Francis, \\ Noriaki Hirose, No\'e P\'erez, Dhruv Shah, Phani T. Singamaneni, Xuesu Xiao, Luis J. Manso%
\thanks{This work involved human subjects. Ethics approval was obtained prior to conducting any research from Aston University's Research Ethics Committee (application nos. EPS21051 and EPS21052).}%
\thanks{Pilar Bachiller-Burgos and Luis V. Calderita are with Universidad de Extremadura, Spain. {\tt\small pilarb@unex.es}}%
\thanks{Luis J. Manso, Ulysses Bernardet and Pranup Chhetri are with Aston University, Birmingham, United Kingdom. {\tt\small l.manso@aston.ac.uk}}%
\thanks{Anthoni Francis is with Logical Robotics, United States of America.}%
\thanks{Noriaki Hirose is with University of California Berkeley and TOYOTA Motor North America, United States of America.}%
\thanks{No\'e P\'erez is with Universidad Pablo de Olavide, Spain.}%
\thanks{Dhruv Shah is with Princeton University, United States of America.}%
\thanks{Phani T. Singamaneni is with LAAS-CNRS, France.}%
\thanks{Xuesu Xiao is with George Mason University, United States of America.}%
}

\usepackage{csquotes}
\usepackage{soul}

\usepackage{multirow}

\usepackage[backend=biber, maxbibnames=99, style=numeric-comp]{biblatex}

\usepackage{tabularx}

\usepackage{amsfonts}
\usepackage{amsmath}
\usepackage{graphicx}

\usepackage{graphicx}
\usepackage{subcaption}

\makeatletter
\def\thickhline{%
  \noalign{\ifnum0=`}\fi\hrule \@height \thickarrayrulewidth \futurelet
   \reserved@a\@xthickhline}
\def\@xthickhline{\ifx\reserved@a\thickhline
               \vskip\doublerulesep
               \vskip-\thickarrayrulewidth
             \fi
      \ifnum0=`{\fi}}
\makeatother

\newlength{\thickarrayrulewidth}
\begin{document}

\maketitle
\thispagestyle{empty}
\pagestyle{empty}

\begin{abstract}
This paper presents a joint effort towards the development of a data-driven Social Robot Navigation metric to facilitate benchmarking and policy optimization for ground robots.
We provide the motivations for our approach and describe our proposal to format and store rated social navigation trajectory datasets.
Following these guidelines,
we compiled a first version of the proposed dataset with 4427 trajectories ---182 real and 4245 simulated--- and presented it to human raters, yielding a total of 4402 rated trajectories after data quality assurance.
Notably, we provide the first all-encompassing learned social robot navigation metric (SN26), along qualitative and quantitative results, including the test loss achieved, a comparison against hand-crafted metrics, and an ablation study.
All data, software, and model weights are publicly available.
\end{abstract}

\section{Introduction}
Social robot navigation is an increasingly relevant field, with both growing academic research interest and notable industry involvement. 
Despite the efforts of the social robot navigation community to establish principles and guidelines for evaluation~\cite{francis2023principles}, benchmarking of social robots is not yet standardized and can be biased to suit specific research narratives.
Social robot navigation (hereafter \textbf{SocNav}) is currently assessed through metrics focusing on success, safety, proxemics, or readability~\cite{mavrogiannis2023core}.
However, different research works use different metrics depending on what their authors consider relevant and often make claims without contemplating tradeoffs between metrics or Pareto optimality.

\begin{figure*}[!t]
    \centering
    \includegraphics[width=0.98\linewidth, clip, trim=0 0 0 0]{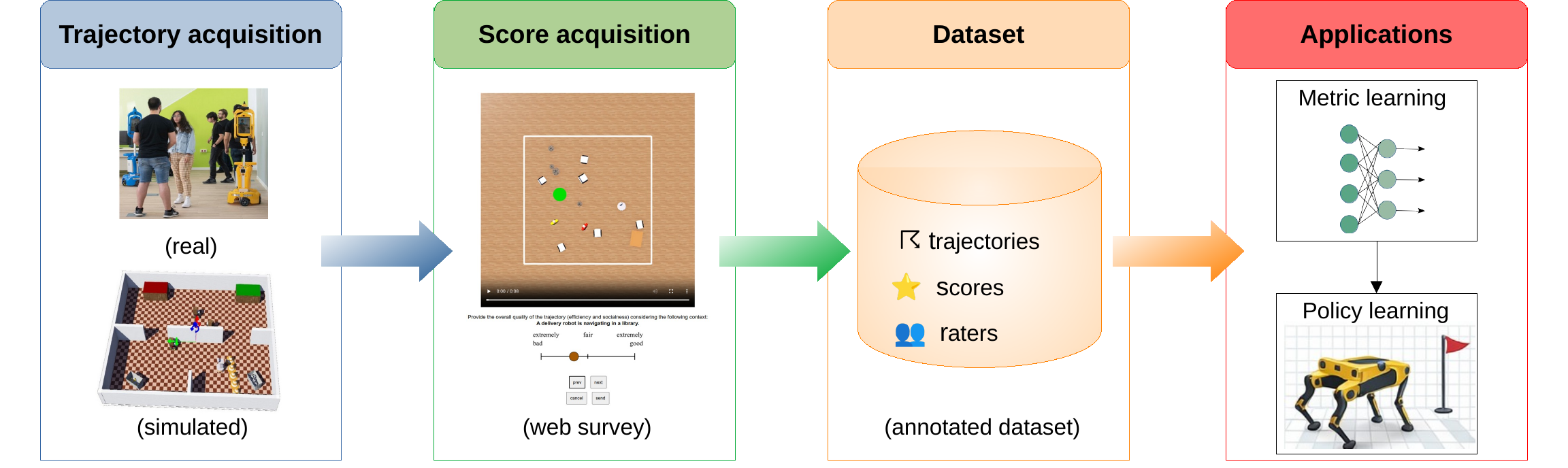}
    \caption{We acquire trajectories from real and simulated scenarios. Simulated trajectories have a lower cost and allow for ethical recording of unsafe behavior, but real trajectories are still required to ensure generalization. We generate top-down videos for the trajectories and show them to raters to collect scores. All data collected is public, including trajectories, raters' demographics, and ratings. While metric and policy learning are use cases for the dataset, the latter is out of the scope of this paper.}
    \label{fig:figure1}
\end{figure*}

\par
Furthermore, task type and navigation context~\cite{francis2023principles} are often overlooked.
How the type of task affects behavior is exemplified by robots entering people's personal space, which is required when a robot delivers an object to a person, but should normally be avoided.
The importance of context is exemplified by whether a robot should pass bystanders at high speed, which is more appropriate when a robot is carrying urgent medical equipment than when returning to base to recharge, even though the task might be the same from a geometric perspective.

\par
Many existing metrics are \textit{analytic}, computed algorithmically from observed variables, which leads to poor variable-wise scalability, and disregard factors such as the type of task or contextual factors such as the distribution of objects around the robot~\cite{francis2023principles}. 
Furthermore, they frequently lack experimental support showing that what the metric measures matches user perceptions of robot behavior quality.
Nevertheless, new metrics continue to be proposed, and, although the intuitions behind them are generally sound, they also often lack empirical support. 
We argue that metrics should ideally be empirically validated, learned from data, and should evaluate the overall effectiveness and socialness of an entire trajectory. We call such a metric an \textbf{A}ll-encompassing \textbf{L}earned \textbf{T}rajectory-wise (\textbf{ALT}) metric~\cite{francis2023principles}. An ALT metric is learned directly from human evaluations of full robot trajectories, conditioned on context ---a category of metrics which, to our knowledge, does not yet exist.

Motivated by the lack of a standardized method to benchmark SocNav policies, limitations of existing analytic metrics, and the absence of an ALT metric, this article studies the feasibility of a data-driven solution to create an ALT metric,  which would solve all the aforementioned problems.

\par
Robots are not yet commonly integrated into everyday human environments with explicit quantitative feedback~\cite{zhang2025predicting}, so no datasets suitable for training learned metrics exist. This gap makes compiling a dataset specifically for this purpose essential. Therefore, this paper contributes with:
\begin{enumerate}
\item a \textbf{specification} for storing social robot navigation trajectories along with their ratings,
\item a proof-of-concept \textbf{dataset} following this specification (\textbf{SocNavData2026}),
\item \textbf{open-source tools} for visualizing the data, and
\item a \textbf{learned metric} exploiting the dataset (\textbf{SN26}).
\end{enumerate}

\par
Our dataset specification includes information concerning the environment, the humans, the robot's task, its context, and the trajectory of the robot, as well as demographic information about the human raters and the scores for the trajectories they rated.
The specification and any dataset that follows it can be exploited for (see Fig.{\ref{fig:figure1}}):
\begin{itemize}
\item \textbf{Training ALT metrics:} Given a sufficiently large and varied set of trajectory-score pairs, a supervised machine learning model can be trained on the dataset to produce a learned ALT metric.
\item \textbf{Policy development:} Given a high-quality ALT metric, SocNav policies can be derived by optimizing the score for the metric ---for example, using reinforcement learning or model predictive control.
\end{itemize}

Our proof-of-concept metric shows that an ALT metric can be learned, and encourages the community to help scale the dataset to the point where metrics learned with it can be used as the \textit{de facto} metric for benchmarking robot trajectories\footnote{Instructions to contribute are in the software repository \url{https://github.com/SocNavData/SocNavData2026}}. Policy development is out of the scope of this paper. %
The experimental results in Sec.~\ref{experimentalresults} provide evidence of the quality of the data compiled and the metric developed.
They include the test loss achieved, an ablation study, an analysis of the correlation between our metric and some of the most common SocNav metrics, and qualitative results for stereotypical scenarios.

\section{Related Work}
Regarding \textbf{learned SocNav metrics}, SocNav1~\cite{manso2020socnav1} and SocNav2~\cite{socnav2_multimedia2022} are the initiatives that most closely resemble the goals of this paper.
SocNav1~\cite{manso2020socnav1} provides scores of static scenarios including one robot and a variable number of humans, with generic object bounding boxes and explicit interactions (human-to-human and human-to-object).
SocNav2~\cite{socnav2_multimedia2022} extended SocNav1 to account for brief periods of time and specific goals, but not full trajectories.
Although these works provide learned SocNav metrics, they do not consider the context of the tasks, and are step-wise, \textit{i.e.}, they account for snapshots rather than whole trajectories.
\par
Although not strictly a metric, another notable research direction
estimates the perturbation of pedestrian movement as a variable to minimize ~\cite{hirose2023sacson, agrawal2024evaluating}.
Despite its elegant simplicity, this approach is dependent on the number of people in the environment, and does not account for the context of the task nor the general perception of trajectory quality.
Similarly, \cite{perez2014robot} learns a cost function over observations of real human motion interactions found in publicly available datasets, then uses that cost to determine a navigation controller. While this cost function could be used as a metric, this approach only considers the context of crossing pedestrians on a pavement and has the same shortcomings as~\cite{hirose2023sacson} and~\cite{agrawal2024evaluating}.

Many \textbf{datasets} relevant to human trajectory modeling exist, but datasets which also include humans, robots and environmental data (such as a map or objects) are scarce.
JRDB~\cite{martin2019jrdb}, THOR~\cite{thorDataset2019}, SCAND~\cite{scanddataset} and SACSoN~\cite{hirose2023sacson} are examples of such datasets.
They provide raw sensor data from teleoperated or autonomous robot trajectories, and frequently include a video feed and 3D LiDAR or RGBD data. However, with the exception of THOR~\cite{thorDataset2019}, pedestrians are annotated as bounding boxes or 2D points \textit{without} orientation. 
Orientation can be very informative as it captures interaction intents~\cite{singamaneni2023survey}, even when humans are static.

\par
We found no datasets containing rated robot trajectories, so we chose to reuse and annotate existing datasets along with recording additional data.
Although THOR includes over 60 minutes of recordings including object annotations, oriented pedestrians and human gaze orientation, we chose to integrate SACSON first because of its scale (75 hours of recordings) and variability (\textit{i.e.}, the trajectories in THOR were all recorded in the same space).
As mentioned in the introduction, we combine synthetic data with real data to vary the quality of the trajectories in order to reduce the bias of the learned metric towards high scores.
In this vein, it must be noted that the new trajectories that we recorded include robot behavior of lower quality than those in existing datasets, contributing to reducing the bias toward higher scores.

\section{Dataset Specification and Tools}
This section first describes our proposed dataset specification, including the variables it represents and its file format, and then summarizes the software we developed to work with this and compatible datasets.

\subsection{Variables}\label{sec:variables}
The dataset includes variables related to raters, trajectories, and rater-trajectory scores.
For every \textbf{rater}, the dataset contains their age (integer), gender (categorical: female, male, non-binary, transgender, other, no-answer), country of residence (categorical), and a rating list.
Each rating list contains tuples $(t, c, r)$, where $t$ is a trajectory identifier (string), $c$ is a context (string), and a $r$ is a \textbf{score} (float, $r \in [0, 1] \subset \mathbb{R}$).
We chose to acquire real-valued scores rather than a Likert scale following best-practice HRI studies~\cite{taylor2022observer} which regard sensitivity as more relevant than repeatability.
To score trajectories, the survey tool used a slider with three references \textit{``extremely bad''} (score $0$), \textit{fair} (score $0.5$) and \textit{``extremely good''} (score $1$).
Note that contexts are not necessarily bound to trajectories when trajectories are recorded; this enables us to vary the context to explore how different contexts affect the perception of the same trajectory.

\par
\textbf{Trajectories} contain data on the robot, the task and its context, humans, objects, and the environment.
Except for variables related to the environment, which apply to the whole trajectory, variables are recorded with a timestamp at each time step.

\par
The \textbf{robot} is described providing its pose (in $SE(2)$), speed (a twist vector), drive system (categorical variable, ``differential'', ``omnidirectional'', ``ackerman'', or ``biomimetic'') and shape (circle, rectangle, or a generic polygon described using a vertex representation).

\par
The \textbf{task} is described by its type, which is a categorical variable (\textit{``go-to''}, \textit{``guide-to''}, \textit{``follow''}, or \textit{``interact-with''}), and its goal.
The \textbf{goal} is specified depending on the type of task.
For \textit{go-to} tasks, the specification requires a target 2D position (in meters) and a target angle (in radians), both with Euclidean and angular distance thresholds, respectively. By varying threshold values, a \textit{go-to} task can specify a very stringent $SE(2)$ point target, just a position, or only an angle.
Tasks of type \textit{guide-to}, are specified similarly to \textit{go-to} tasks, but require an additional human identifier, and the task target applies to the human, not the robot.
Tasks of type \textit{follow} and \textit{interact-with} require a human identifier to specify which human is the robot expected to follow, along with an Euclidean distance threshold.

\par
\textbf{Humans} are described with a numerical identifier, their 2D pose and, optionally, the 3D coordinates of their COCO-18 key-points. The focus group decided making the 3D version of the human pose mandatory would limit adoption, as few research groups or companies have access to full human pose trackers where robots operate. However, the focus group decided to add 3D human pose as an optional element so the specification would not become prematurely obsolete. 

\par
\textbf{Objects} are described with an identifier, a text-based description, their 2D pose and their shape (circle, rectangle, or polygon). The focus group chose a text-based description over a predefined set of object types because of its flexibility (text enables open-set objects) and fidelity (language models can often easily turn text into categories, but the reverse is not true without loss of information).

\par
The remaining \textbf{environmental} information consists of the walls (specified as 2D polylines), a 2D grid map (with the same fields as ROS2 maps), and area semantics (text-based, following the same rationale as the text-based object description).
The grid map is specified with a 2D data array, and the resolution of the grid is specified in metres per cell.

\subsection{Structure and storage}
The raw dataset contains two main directories: one containing the \textit{trajectories} and another one containing data about the raters and their \textit{ratings}.
The \textit{trajectories} directory contains JSON files of recorded trajectories in sub-directories named according to the source of the trajectory data. Each file contains one trajectory, and their relative file paths serve as trajectory identifiers.
The \textit{ratings} directory contains a separate JSON file for each questionnaire completed. 
In addition to the rater's demographic information, each of these files store their ratings as a list of tuples: trajectory identifier (\textit{i.e.}, a relative path to the JSON file), context, and score, as described in~\ref{sec:variables}.

\subsection{Software layer}
In the Git repository of the project\footnote{\url{https://github.com/SocNavData/SocNavData2026}}, we provide scripts to visualize and perform statistical analyses on the data, classes for data-loading, data augmentation and normalization, and code to deploy the surveying tool to acquire ratings.

\textbf{Statistical analysis scripts:} Used to compute the quadratic weighted Cohen's kappa coefficient of the raters and to produce the consistency map in Fig.~\ref{fig:consistency-matrix}. More details are provided in Sec.~\ref{sec:dataquality}.

\textbf{Trajectory visualization tool:}
Unlike other types of datasets with simpler datatypes, such as images, or audio, SocNav data combines different variables, which makes errors likely if the trajectories are not carefully compiled.
Inconsistencies in coordinate or unit systems are common issues that can be picked up with the tools provided.

\textbf{Surveying tool:} Ratings surveys benefit from usable ---and aesthetic--- representations. The surveying tool shows video recordings along with randomly generated contexts, and asks raters to provide their scores.

\textbf{Data normalization, and augmentation:} 
Normalization and augmentation help to include equivariances and invariances in machine learning models without embedding them into equations.
To avoid forcing users to duplicate complex and error-prone data augmentation code, and to streamline and homogenize data augmentation and normalization, data transforms are provided.
Data normalization automatically reframes all poses into the goal frame of reference; this can steer machine learning models towards invariance \textit{w.r.t.} the frame of reference.
The provided software implements three types of data augmentations.
First, to enhance robustness to noise, poses can be augmented with additive Gaussian noise.
Secondly, when goal orientations exceed a threshold larger or equal to $\pi$~rad, the orientation value is randomized.
Finally, trajectory data can be randomly mirrored (left-right).



\section{Description and Analysis of the Dataset}\label{datasets}
We provide a first dataset version (SocNavData2026) to show how the proposed dataset specification can be exploited to generate a learned metric.
We collected $4427$ trajectories, $182$ real and $4245$ simulated. 
We extended this initial set by varying the configuration of the walls in the scenarios of the compiled trajectories (\textit{i.e.}, removing some or all of the walls).
This process resulted in a total of $7472$ trajectories. It is key to have both real and simulated data sources because they pose different challenges: real trajectories typically have lower variability than simulated ones because of the expense of recording trajectories in multiple scenarios, while simulated data have less realistic movements and noise.
Additionally, due to safety and ethical reasons, dangerous robot trajectories can only be acquired in simulation.
Furthermore, acquiring a large enough dataset to cover all possible situations with real data exclusively would be unrealistically expensive.

\par
To validate the feasibility of the approach and ensure coverage of practically relevant scenarios, we implemented a focused sampling strategy for this first version.
We selected the domain of indoor areas whose bounding box sides range between $6~m$ and $10~m$.
The navigation tasks covered in the dataset are all of the type \textit{``go-to''}; other tasks such as following or guiding humans are left for future additions to the dataset repository.
The objects considered in the dataset are limited to chairs, shelves, tables, lights, and plants.
The maximum robot speed recorded is $2~m/s$.

\subsection{Trajectories recorded}
Two sources were used to acquire \textbf{real trajectories}.
A set of 109 trajectories was recorded at the ARP laboratory (Aston University).
A $8\times7.4$ meters wide room was populated with tables, shelves, chairs, a TV, and from 1 to 6 humans.
Another 73 trajectories were re-used from the SACSoN dataset~\cite{hirose2023sacson}.
The trajectories available in SACSoN were recorded in environments larger than the maximum area set for the dataset, so they were cropped to comply with the range allowed within the restrictions for the dataset.
We included \textbf{simulated trajectories} from a range of simulators and algorithms, using both manual control and a learned RL agent on SocNavGym~\cite{kapoor2023socnavgym}, HATEB~\cite{singamaneni2021human}, and HuNavSim~\cite{perez2023hunavsim}, as well as manual control on Gazebo.
Ethics approval was obtained prior to conducting any research from Aston University's Research Ethics Committee (application numbers EPS21051 and EPS21052).

\subsection{Data quality}\label{sec:dataquality}
Each survey included $15$ control questions to assess consistency between the participants. 
Additionally, five of these questions were presented twice to evaluate intra-rater consistency. 
In total, $49$ participants scored $6481$ trajectories. 

\par
To ensure data quality, we first discarded ratings from participants who did not answer all control questions, as it would not be possible to analyze their consistency. 
For the $34$ remaining participants, we applied a selection process that considers both intra-rater and inter-rater consistency using the quadratic weighted Cohen's kappa coefficient.
We define $\mu_{intraH}$ as a high-quality threshold and $\mu_{intraL}$ as a minimum-quality threshold for intra-rater consistency and $\mu_{inter}$ as a minimum threshold for inter-rater consistency. 
The selection consists of the following steps:
\begin{enumerate}
    \item Compute the average score for each control question over the population of raters with an intra-rater consistency greater than $\mu_{intraH}$.
    \item Discard raters with an intra-rater consistency lower than $\mu_{intraL}$ or an inter-rater consistency lower than $\mu_{inter}$ \textit{w.r.t.} the average scores of the control questions. 
\end{enumerate}

\par
\begin{figure}[!t]
    \centering
    \includegraphics[width=0.7\columnwidth, clip, trim=0 0 0 0]{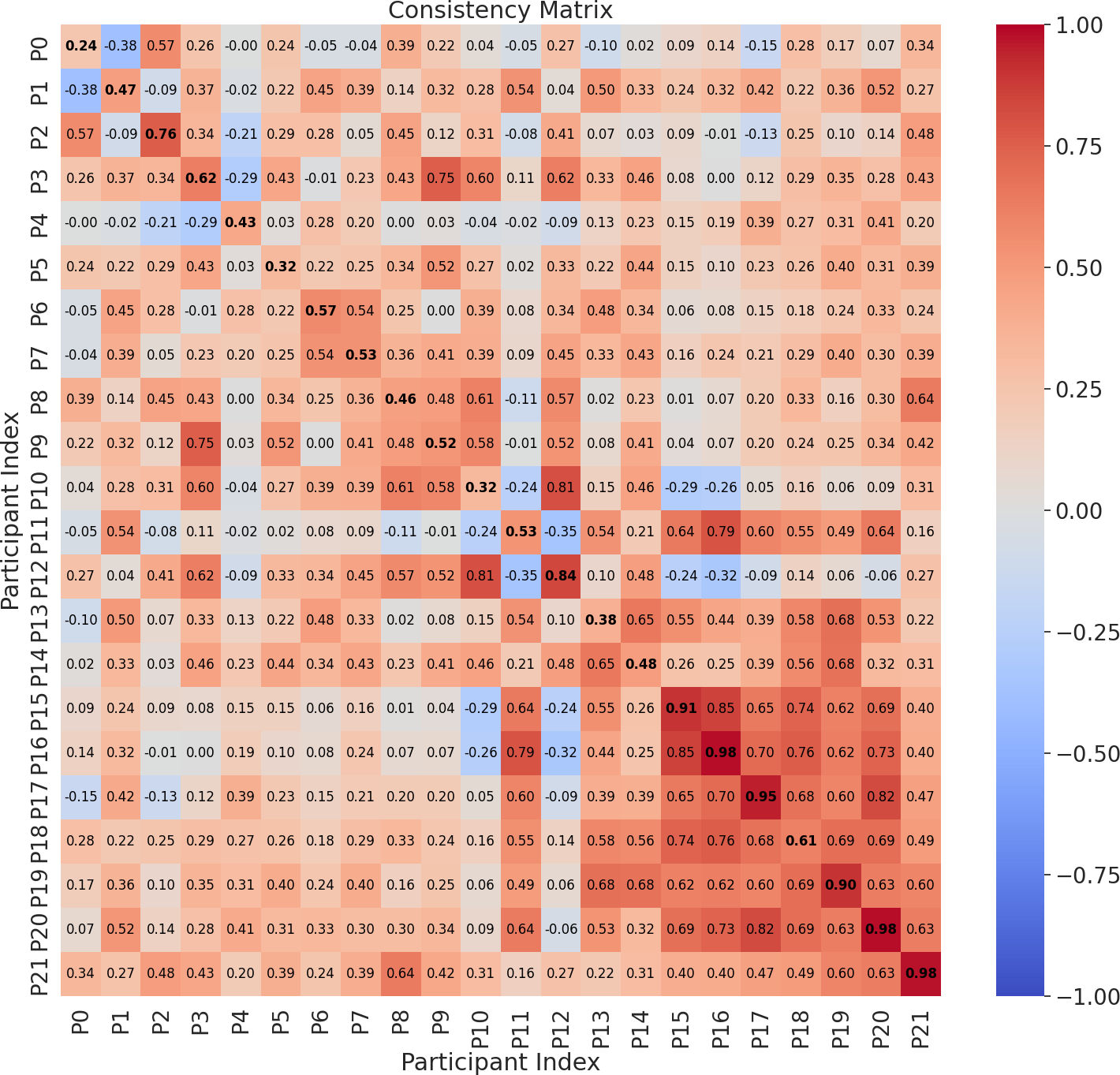}
    \caption{Consistency map of selected raters.}
    \label{fig:consistency-matrix}
\end{figure}

Due to the limited number of repeated control questions, small variations in the answers of a rater could have a significant impact on their intra-rater agreement~\cite{temel2017samplesize}. 
Thus, an initial selection based only on high intra-rater consistency could result in discarding too many raters.
Using both coefficients, we retain raters who generally align with the group opinion, even with minor inconsistencies in their responses. 
This approach improves the overall reliability and representativeness of the data by filtering out erratic and atypical ratings without being excessively restrictive.
Figure~\ref{fig:consistency-matrix} shows the consistency map of the raters selected following the described procedure, considering values of $0.4$, $0.1$ and $0.2$ for $\mu_{intraH}$, $\mu_{intraL}$ and $\mu_{inter}$, respectively. 
The responses of 8 raters were discarded due to very low intra-rater consistency.
Additionally, 6 raters presented significant deviations from the mean opinion according to $\mu_{inter}$.
The resulting set of \textbf{22 raters} yields a \textbf{final dataset of $\textbf{4402}$ scored trajectories}.

\section{Experimental Results}\label{experimentalresults}
We propose a Gated Recurrent Unit-based neural network (GRU~\cite{cho2014learning}) ALT metric model to demonstrate the feasibility of data-driven approaches and to serve as a baseline in future work.
This section describes the data transformations applied, the architecture used, the training procedure, and the quantitative and qualitative results.

\subsection{Data transformations}
The raw trajectory data is augmented using mirroring and transformed into 1D vectors that serve as input to the GRU.
The vectors ($x_0, \dots, x_t, \dots, x_T$) are the concatenation of trajectory features ($f_t$),  metric-based features ($m_t$), and an ad-hoc context embedding ($c_t$): $x_t = f_t || m_t || c_t$

\par
The trajectory features, $f_t$, include pose, speed, and acceleration of the robot \textit{w.r.t.} the goal frame of reference, goal thresholds (regarding position and orientation), general scenario information (flags indicating the presence of humans and the existence of walls in the scenario), and data that allows contextualizing in time the current step in the whole trajectory (the current step ratio and a last step flag).
The metrics compiled in $m_t$ are the following:
\begin{itemize}
    \item A flag indicating whether the robot reaches the goal.
    \item Distances to the nearest human, object, and wall.
    \item Flags indicating whether the robot has collided with a human, an object, or a wall.
    \item Number of humans within a radius of the robot ($0.4$, $0.6$, and $0.8~m$).
    \item Flags indicating social space intrusions, considering the same thresholds as in the previous point.
    \item Features indicating potential danger to humans: minimum time to collision, maximum \textit{cost of fear}, and maximum \textit{cost of panic} as defined in~\cite{singamaneni2023towards}.
    \item Minimum distance from humans up to the current step.
    \item Path efficiency ratio measured as the initial distance to the goal divided by the length of the trajectory up to the current step.
\end{itemize}

\par
The contextual values in $c_t$ are computed from the free-form text describing the context of the task querying an LLM (Anthropic's Claude 3.7 Sonnet)\footnote{The model was trained using the quantization generated by Claude 3.7 Sonnet v20250219, Prometheus v2.0, gpt-oss-20b, LLaMa 3.3 and Deepseek R1. Claude produced the best model loss.}.
The prompt used was: \textit{``I will give a task description for a robot. I want you to reply with the percentile (a number from 0 to 100) that corresponds to {\underline{\textless{}VARIABLE\textgreater}} in comparison with that of other tasks you could imagine. I don't want an explanation, only the percentile. Take your time to think, but respond with a single integer from 0 to 100.''}, where \textit{\textless{}VARIABLE\textgreater} is substituted with each of the strings in Table~\ref{tbl:contextAspects}.
The output of the LLM is later normalized to the range 0-1.

\begin{table}[h]
\caption{Contextual queries used in the LLM prompts.}
\centering
    \begin{tabularx}{0.9\columnwidth}{X}
        \specialrule{0.7pt}{0em}{0.1em} 
          \textbf{Contextual query} \\
        \specialrule{0.7pt}{0.1em}{0.1em} 
          the urgency of the task
       \\ the importance of the task
       \\ the risk involved in the task
       \\ the distance the robot should keep to humans during the task
       \\ the distance the robot should keep to objects during the task
       \\ the speed with which the robot should move
       \\ the importance of comfort versus efficiency in the task
       \\ to what extent bumping into a human would be justified
       \\ to what extent bumping into an object would be justified
       \\ the importance of moving in a predictable way
       \\
       \specialrule{0.7pt}{0.1em}{0em} 
    \end{tabularx}
  \label{tbl:contextAspects}
\end{table}


\subsection{Training procedure}
Discarding the trajectories used as control questions, we create a $0.9$/$0.05$/$0.05$ train/validation/test split and train a GRU-based neural network.
The split is generated randomly constraining all variants of each trajectory to be assigned to the same subset to prevent data leakage.
The split ratio was chosen to maximize the amount of data available for training, given the diversity of contexts and trajectories.
Considering the dataset’s moderate size, it ensures stable learning while keeping the validation and test sets sufficient to assess generalization.
In total, these subsets correspond to $181$ and $211$ trajectories, $44$ and $50$ minutes worth of trajectory data, respectively, which we consider sufficient for evaluation.
\par
The network consists of a GRU connected to an MLP. The GRU has 4 layers with 256-unit hidden layers, and the output of the last step of its sequence is fed into an MLP with a 256-unit hidden layer and a 1-unit output layer.
The hidden layers use LeakyReLU activation with a $0.01$ negative slope, and the output layer uses sigmoid activation.
The model is trained with a batch size of $32$ and a learning rate of $1e-5$, using MSE loss with the target of the raters' scores.
The training follows an early-stopping strategy with a patience of $20$ epochs.

\subsection{Quantitative results}
\begin{figure}[!t]
    \centering
    \includegraphics[width=0.85\columnwidth, clip, trim=0 0 0 0]{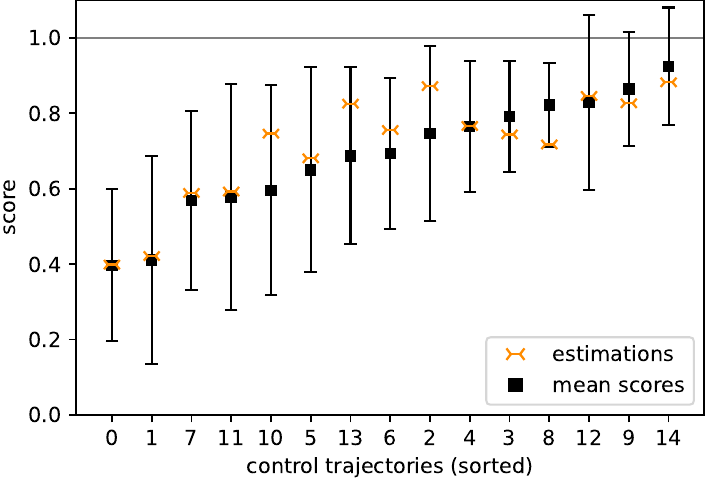}
    \caption{Plot for the control questions with their corresponding mean, standard deviation, and the estimations made by the model (SN26 metric). The control questions are shown sorted according to their mean score.}
    \label{fig:control-results}
\end{figure}

We evaluated the model’s performance on the test set, obtaining a final MSE of $\mathbf{0.0457}$ and a corresponding MAE of $\mathbf{0.160}$.
In addition, we analyzed the model’s behavior on the control questions. 
Since these questions were rated multiple times, their mean score provides a reliable reference, allowing us to quantify the model’s deviation from the average human response.
Fig.~\ref{fig:control-results} shows the mean scores for the control questions, along with their standard deviations, and the estimation the model made for them.
The MSE and MAE of the model \textit{w.r.t.} the mean of the control questions were $0.0053$ and $0.053$, respectively.

We used Pearson (\textit{r}) and Spearman (\textit{p}) correlations to measure how common analytic metrics and the learned metric proposed align with human scores.
The results, in Table~\ref{tab:corr_scores}, show a much higher alignment between human scores and the proposed ALT metric.
Metrics related to proxemics (social space intrusions ratio, and maximum number of near humans) show a certain level of alignment, but it can be considered low-moderate.

\begin{table}[!t]
\caption{Pearson (\textit{r}) and Spearman (\textit{p}) correlations with humans' scores for different SocNav metrics and the learned one on the test set. (A), (B) and (C) refer to different distance thresholds: $0.4$, $0.6$, and $0.8$. 
Results yielding p-values greater than $0.05$ have been considered to have low statistical significance and marked in red.
}
\centering
    \begin{tabularx}{0.8\columnwidth}{lcc}
        \specialrule{0.7pt}{0em}{0.1em} 
        & \textit{r} & \textit{p} \\
        \specialrule{0.7pt}{0.1em}{0.1em} 
success &0.366  &0.280  \\
min. distance to humans & {\color{darkred}0.009}  &0.397 \\
min. distance to objects &{\color{darkred}-0.006} &{\color{darkred}0.100} \\
collisions with humans &-0.385  &-0.374  \\
collisions with objects &-0.387 &-0.365  \\
social space intrusions ratio (A) &-0.316 &\underline{-0.422} \\
social space intrusions ratio (B) &-0.271 &-0.341  \\
social space intrusions ratio (C) &-0.184 &-0.252 \\
max. num. of near humans (A) &\underline{-0.415}&-0.420  \\
max. num. of near humans (B) &-0.303 &-0.328 \\
max. num. of near humans (C) &-0.253  &-0.283 \\
min. time to collision &{\color{darkred}0.121} &0.313  \\
max. fear &-0.341 &-0.303 \\
maximum panic &{\color{darkred}-0.128} &-0.156 \\
path efficiency ratio &{\color{darkred}0.105} &{\color{darkred}0.050} \\
cumulative heading changes &{\color{darkred}0.049} &{\color{darkred}0.035} \\
\textbf{learned metric (ours)} & \textbf{0.797} & \textbf{0.684} \\
       \specialrule{0.7pt}{0.1em}{0em} 
\end{tabularx}%
\label{tab:corr_scores}
\end{table}

To analyze the influence of different groups of input features on the performance of the model, an ablation study was carried out.
Table~\ref{tab:ablation} reports the effect of removing groups of input features using the MAE, and the Concordance Correlation Coefficient (CCC) to quantify such effect. 
The first column corresponds to the original model, which uses the complete set of features.
Values indicating a greater deterioration in the metrics have been boxed.
Results show that the most critical sets of features are \textit{Collision} and \textit{Intrusion}, which include the collision flags and the social space intrusion variables.
Other important features are those in \textit{Success}, which indicate the task's completeness.
On the contrary, context features in \textit{Urgency} (task urgency and robot speed) have a limited influence on the output of the model.
This can be explained by certain inconsistencies introduced by the LLM, among some context variables. 
Such inconsistencies are expected to diminish as LLMs continue to improve in accuracy and consistency.

\begin{table*}[!h]
\caption{Ablation study (MAE and CCC) for the full model and feature-group removals. Columns report performance after removing the corresponding feature group.}
\label{tab:ablation}

\centering
\begingroup
\renewcommand{\tabcolsep}{8.5pt}
\renewcommand{\arraystretch}{0.95}

\begin{threeparttable}

 {
 
\begin{tabular}{lcccccc}   

\specialrule{0.7pt}{0.1em}{0.1em}
\multicolumn{1}{c}{} & \multicolumn{1}{c}{} & \multicolumn{5}{c}{\textbf{Robot \& Task}}  \\
\cmidrule(lr){3-7}
& \textbf{All features} & Scenario & Pose & Speed & Success & Path  \\
\specialrule{0.7pt}{0.1em}{0.1em}
MAE&\textbf{0.160}&$0.183$&$0.167$&$0.165$&$0.190$& $0.165$  \\
CCC&$\mathbf{0.776}$&$0.700$&$0.753$&$0.760$&$0.681$&$ 0.757$ \\
\specialrule{0.7pt}{0.1em}{0.1em}
\multicolumn{1}{c}{} & \multicolumn{1}{c}{} & \multicolumn{5}{c}{\textbf{Social \& Safety}} \\
\cmidrule(lr){3-7}
& \textbf{All features} & Distance & Collision & Intrusion & Near & Danger \\
\specialrule{0.7pt}{0.1em}{0.1em}
MAE&\textbf{0.160}& $0.173$&$0.190$&$\boxed{0.191}$&$0.165$&$0.164$ \\
CCC&$\mathbf{0.776}$&$0.726$&$\boxed{0.650}$&$0.652$&$0.755$&$0.754$ \\
\specialrule{0.7pt}{0.1em}{0.1em}
\multicolumn{1}{c}{} & \multicolumn{1}{c}{} &  \multicolumn{1}{c}{} & \multicolumn{3}{c}{\textbf{Context}} & \\
\cmidrule(lr){4-6}
&\textbf{All features} & &Urgency & Risk & Importance & \\
\specialrule{0.7pt}{0.1em}{0.1em}
MAE&\textbf{0.160}&& $0.162$&$0.161$&$0.165$& \\
CCC&$\mathbf{0.776}$&&$0.774$&$0.768$&$0.742$& \\

\specialrule{0.6pt}{0.05em}{0em}
\end{tabular}
}

\begin{tablenotes}[flushleft]
\item \textit{Ablations:} \textit{\textbf{Scenario}} = goal thresholds and existence flags; \textit{\textbf{Pose}} = robot pose; \textit{\textbf{Speed}} = robot speed and acceleration; \textit{\textbf{Success}} = success flag and distance to goal; \textit{\textbf{Path}} = path efficiency ratio, step ratio, episode end; \textit{\textbf{Distance}} = distances to humans, objects, and walls; \textit{\textbf{Collision}} = collisions with humans, objects, and walls; \textit{\textbf{Intrusion}} = social space intrusions; \textit{\textbf{Near}} = number of near humans; \textit{\textbf{Danger}} = minimum TTC, maximum panic, maximum fear; \textit{\textbf{Urgency}} = task urgency and preferred robot speed; \textit{\textbf{Risk}} = task risk, distances and justification of bumping into humans and objects; \textit{\textbf{Importance}} = task importance, human comfort, robot predictability.

\end{tablenotes}

\end{threeparttable}
\endgroup
\end{table*}

\subsection{Qualitative results}\label{qualitative}

\begin{figure*}[tbh]
\centering
\begin{minipage}{0.48\textwidth}
  \centering
  \includegraphics[width=0.99\columnwidth, clip, trim=0 0 0 0]{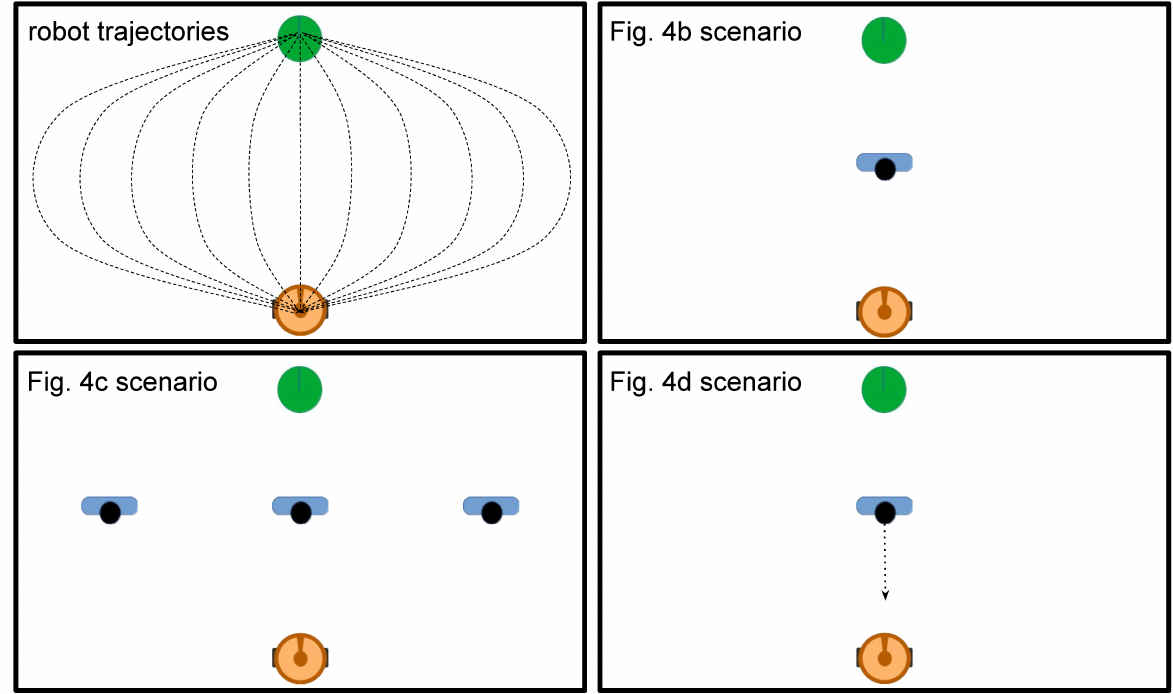}\\
  {\scriptsize\textbf{(a)} Robot trajectories and scenarios.}
\end{minipage}
\hfill
\begin{minipage}{0.48\textwidth}
  \centering
  \includegraphics[width=0.99\columnwidth, clip, trim=0 0 0 0]{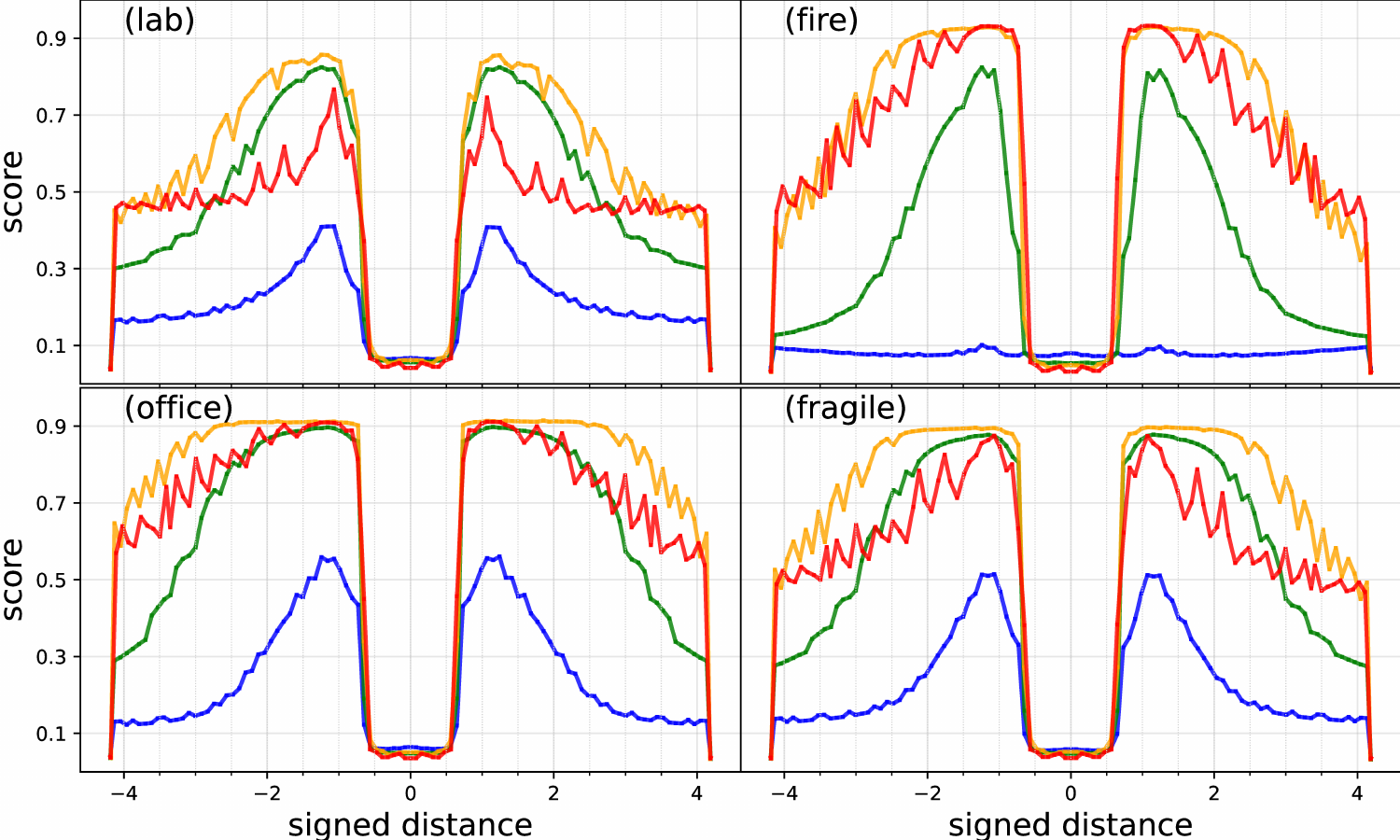}\\
  {\scriptsize\textbf{(b)} One static person.}
\end{minipage}
\\\vspace{1.8mm}
\begin{minipage}{0.48\textwidth}
  \centering
  \includegraphics[width=0.99\columnwidth, clip, trim=0 0 0 0]{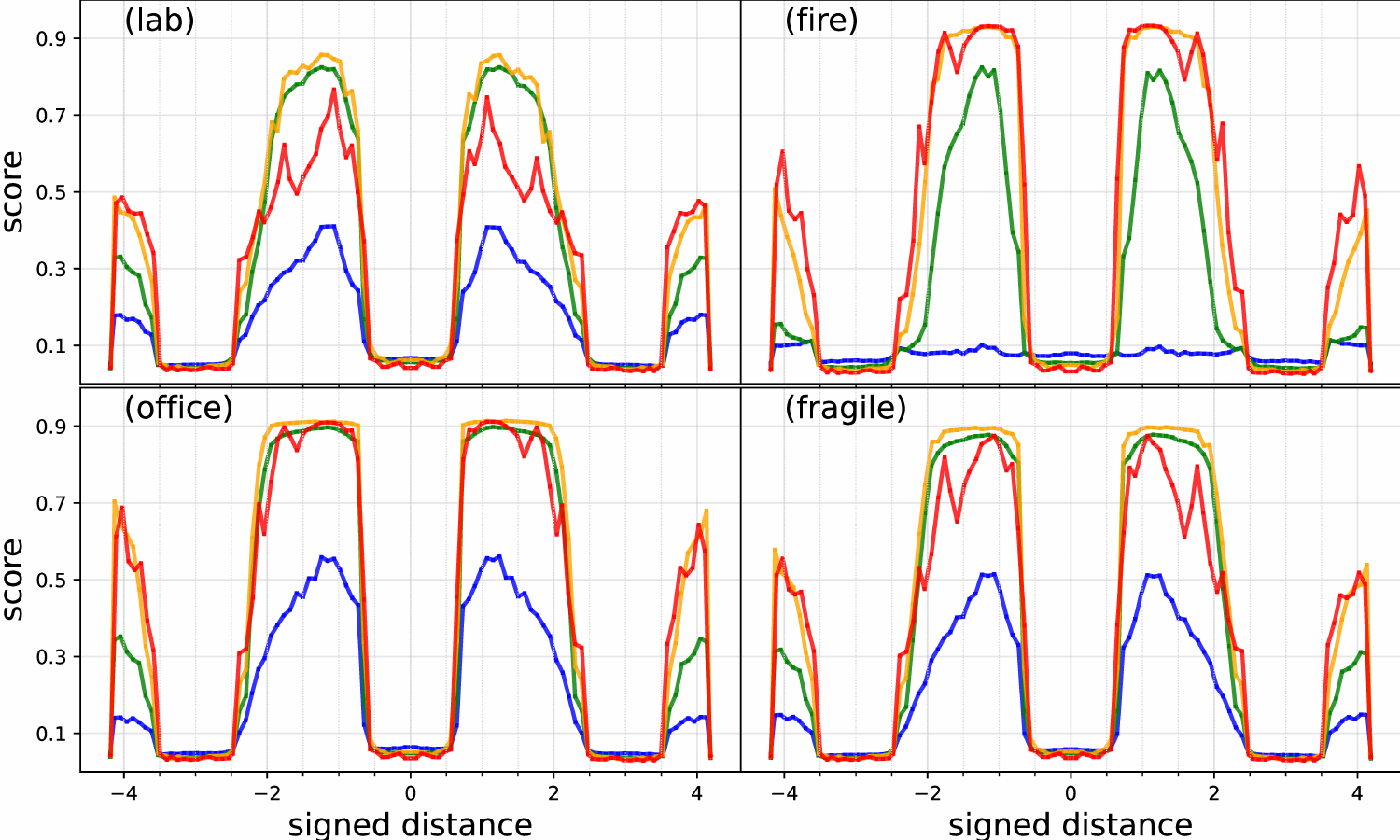}\\
  {\scriptsize\textbf{(c)} Three static persons.}
\end{minipage}
\hfill
\begin{minipage}{0.48\textwidth}
  \centering
  \includegraphics[width=0.99\columnwidth, clip, trim=0 0 0 0]{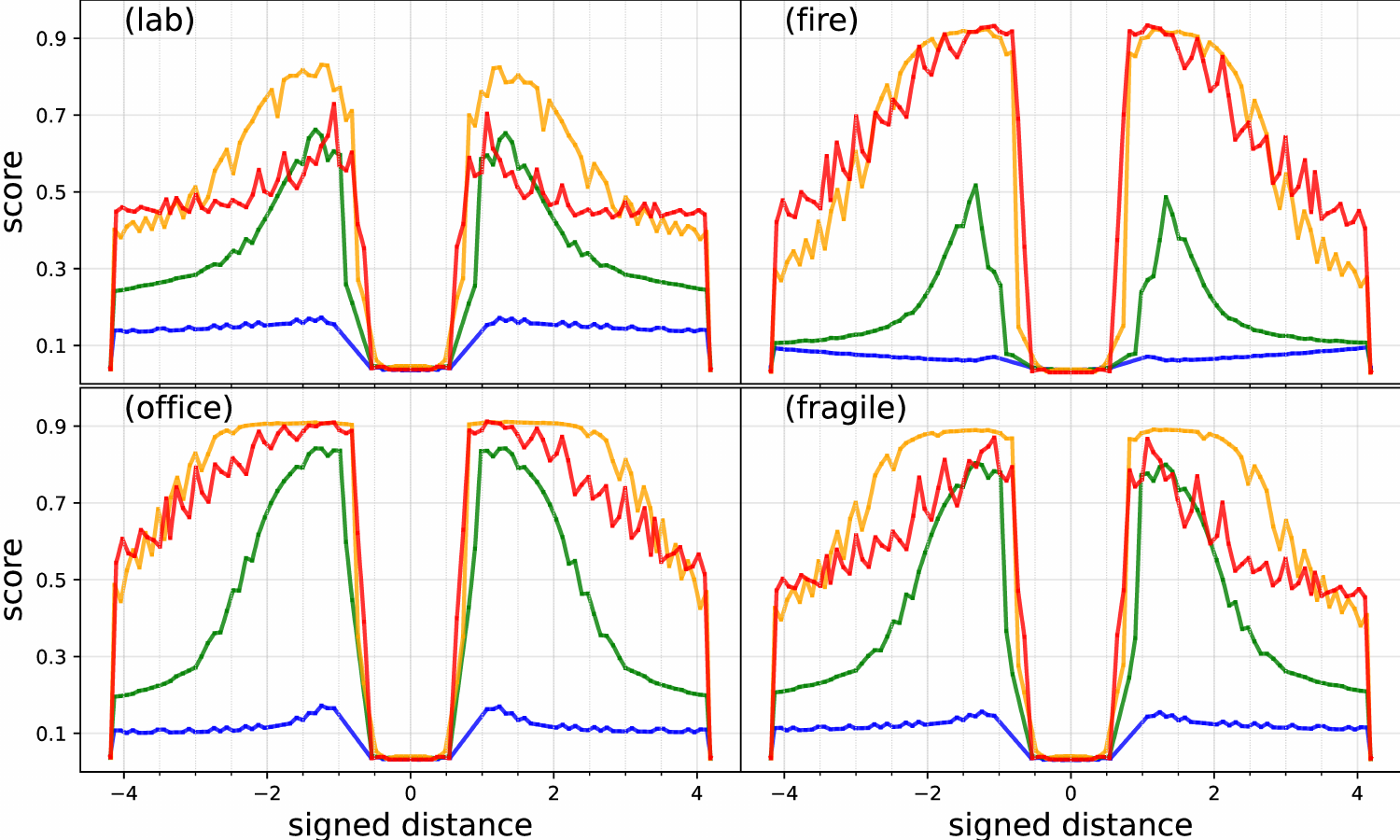}\\
  {\scriptsize\textbf{(d)} A person moving in the robot direction.}
\end{minipage}
\caption{Output of the learned metric (Figs.~\ref{fig:qual_results_rnn}(b)--(d)) for different trajectory variations and evaluation scenarios (illustrated in Fig.~\ref{fig:qual_results_rnn}(a)), contexts, and speeds.
To reach its goal, the robot follows different trajectories, with varying degrees of divergence from the central line (top-left image of Fig.~\ref{fig:qual_results_rnn}(a)).
The horizontal axis of Figs.~\ref{fig:qual_results_rnn}(b)--(d)) refers to this divergence, with each value representing a trajectory.
The context identifiers are shown in the top-left corner of each subplot and described in Sec.~\ref{qualitative}.
Speed is color-coded:
\textcolor{myblue}{\raisebox{0.3ex}{\rule[0.1ex]{0.4cm}{2pt}}}~0.20~m/s
\textcolor{mygreen}{\raisebox{0.3ex}{\rule[0.1ex]{0.4cm}{2pt}}}~0.40~m/s
\textcolor{myorange}{\raisebox{0.3ex}{\rule[0.1ex]{0.4cm}{2pt}}}~0.80~m/s
\textcolor{myred}{\raisebox{0.3ex}{\rule[0.1ex]{0.4cm}{2pt}}}~1.60~m/s.
}

\label{fig:qual_results_rnn}
\end{figure*}

To test the model from a qualitative perspective, we generated three additional simulated scenarios where people are located frontally between the robot's initial position and the goal position (Fig.~\ref{fig:qual_results_rnn}(a)). 
The robot (in orange) moves at different speeds following several trajectories (a total of $101$) with varying lateral deviations from the straight line to the goal (green circle).
The top left image in Fig.~\ref{fig:qual_results_rnn}(a) shows a subset of these trajectories as dotted curves.
A top view of the three scenarios is shown in the remaining images of Fig.~\ref{fig:qual_results_rnn}(a). In the two scenarios on the right of Fig.~\ref{fig:qual_results_rnn}(a), a single human (in blue) is present: in the first case static, and in the other approaching the robot's initial location. The third scenario corresponds to three static humans positioned in front of the robot.
Additionally, four different contexts are considered for each scenario, each with distinct urgency, importance, and risk among other variables: \textit{``A robot is working with lab samples. The samples contain a deadly virus''} (lab);
\textit{``A restaurant robot is looking for a fire extinguisher, as it just detected a fire''} (fire);
\textit{``An office assistant robot keeps track of who is in the office today''} (office); and
\textit{``A delivery robot is navigating in a hospital. It works with fragile objects''} (fragile).

\par
Figures~\ref{fig:qual_results_rnn}(b),~\ref{fig:qual_results_rnn}(c), and ~\ref{fig:qual_results_rnn}(d) show the score assigned by the model to each trajectory for these three scenarios.
The horizontal axis represents the maximum deviation of the trajectory from the straight line ---the sign refers to the side, left or right.
As these figures show, the model learns to distinguish among contexts.
The highest speed is penalized more in the contexts labeled as \textit{lab} and \textit{fragile}, where the robot would generally be expected to move slowly due to the risk of the task.
In these two contexts, the robot is also expected to maintain a wider distance from humans, which explains why trajectories closer to the human (\textit{i.e.}, with smaller deviations) receive lower scores than in the other contexts.
In contrast, the \textit{fire} context reflects a high-urgency situation: the lowest speed results in scores close to zero on all trajectories and scenarios, and closer proximity to humans is more acceptable.
\par
In addition, in the \textit{office} context, where the task is less critical, the lowest speed is less penalized and the scores decay more slowly than in the other contexts as the robot deviates from the straight line to the goal.
In the scenario with three pedestrians (Fig.~\ref{fig:qual_results_rnn}(c)), this behavior is less evident.
Nevertheless, it can be observed how, for the range of distances with the highest scores, the \textit{office} context receives practically the same score, while the rest of the contexts are penalized as the deviation increases.
\par
Finally, notable differences can be observed between the two scenarios with one person (Fig.~\ref{fig:qual_results_rnn}(b) and (d)).
When the person moves, the robot should move faster than when the person is static.
This aligns with the expectation that the robot should actively maneuver in time to avoid the moving person safely.
Additionally, deviating farther from the straight line is more penalized than in the static scenario, which can be explained by the fact that, once the robot has safely avoided the human, the distance from them is greater than in the stationary person scenario, as the person moves away from the robot.

Overall, these qualitative results show that the model not only distinguishes among contexts but also captures the balance between speed, proxemics, and urgency, and that it aligns with human expectations of safe and efficient behavior.

\section{Conclusions and Future Work}
This paper described how All-encompassing Learned Task-wise metrics~\cite{francis2023principles} can be learned following a data-driven approach in the context of social robot navigation.
The experimental results support the idea that the approach is feasible and that the SN26 metric correlates with raters' scores better than common analytic metrics.
The quantitative results achieve MSE and MAE of $0.0457$ and $0.160$, respectively.
The qualitative results show that the learned metric is sensitive to both the content of the scenarios and the context of the tasks.

\par
The authors acknowledge these results could be enhanced with additional data and architectural improvements, which points to the need for further research in this area.
More sophisticated architectures were not satisfactory in our experiments, arguably due to the size of the dataset.

\par
Our envisioned future work spans two different areas: a)~increasing the dataset size with new trajectories and trajectories from existing non-rated datasets~\cite{thorDataset2019,martin2019jrdb}, and b)~development of domain-specific deep learning architectures that can more efficiently leverage geometric relationships.


\printbibliography

\end{document}